\documentclass[runningheads]{llncs}
\usepackage[T1]{fontenc}
\usepackage{amsmath,amssymb}
\usepackage{subcaption}
\usepackage{algorithm}
\usepackage{algpseudocode}
\usepackage{graphicx,verbatim}
\begin{document}
\title{Disentanglement of Biological and Technical Factors via Latent Space Rotation in Clinical Imaging Improves Disease Pattern Discovery}

%

\author{Jeanny Pan\inst{1,2}, Philipp Seeb\"ock\inst{1,2}, Christoph F\"urb\"ock\inst{1,2}, Svitlana Pochepnia\inst{3}, Jennifer Straub\inst{1,2}, Lucian Beer\inst{3}, Helmut Prosch\inst{3}, Georg Langs\inst{1,2}}  
\authorrunning{Pan et al.}
\titlerunning{Latent Space Rotation Disentanglement}

\institute{
\textsuperscript{1}Computational Imaging Research Lab, Department for Biomedical Imaging and Image-guided Therapy\\
\textsuperscript{2}Comprehensive Center for Artificial Intelligence in Medicine\\
\textsuperscript{3}Division of General and Pediatric Radiology, Department for Biomedical Imaging and Image-guided Therapy\\
Medical University of Vienna\\
\email{jeanny.pan@meduniwien.ac.at, georg.langs@meduniwien.ac.at}
}

\maketitle              
\begin{abstract}

Identifying new disease-related patterns in medical imaging data with the help of machine learning enlarges the vocabulary of recognizable findings. This supports diagnostic and prognostic assessment. However, image appearance varies not only due to biological differences, but also due to imaging technology linked to vendors, scanning- or reconstruction parameters. The resulting domain shifts impedes data representation learning strategies and the discovery of biologically meaningful cluster appearances. To address these challenges, we introduce an approach to actively learn the domain shift via \emph{post-hoc} rotation of the data latent space, enabling disentanglement of biological and technical factors.   
Results on real-world heterogeneous clinical data showcase that the learned disentangled representation leads to stable clusters representing tissue-types across different acquisition settings. Cluster consistency is improved by $+19.01\%$ (ARI), $+16.85\%$ (NMI), and $+12.39\%$ (Dice) compared to the entangled representation, outperforming four state-of-the-art harmonization methods. When using the clusters to quantify tissue composition on idiopathic pulmonary fibrosis patients, the learned profiles enhance Cox survival prediction. This indicates that the proposed label-free framework facilitates biomarker discovery in multi-center routine imaging data. Code is available on GitHub \url{https://github.com/cirmuw/latent-space-rotation-disentanglement}.


\end{abstract}

\section{Introduction}\label{sec:introduction}
Unsupervised machine learning can identify imaging patterns associated with disease or its progression. It can expand the vocabulary of markers for assessing and predicting disease course~\cite{Pan2023-qy,Lee2024-nu}. However, in clinical practice, the heterogeneity of imaging protocols and reconstruction parameters introduces variability that hinders the replicable identification of appearance patterns across real world routine patients~\cite{Yoon2024,Zhou2021}.
In computed tomography (CT), factors including manufacturer, reconstruction kernels, or slice thickness can severely limit the comparability of quantitative analyses across institutions and in routine data~\cite{Mets2012-ao,Gierada2010-ik,Boedeker2004-jx,Prayer2021-qv}. For instance, outcomes significantly varied in aggressive non-Hodgkin’s lymphoma due to unrecognized heterogeneity across multiple centers~\cite{Fisher2004-tz}. This has prompted growing research efforts toward algorithms that mitigate the effects of technical heterogeneity.

\paragraph{Related work in the area of disentanglement.}
Early latent-factor methods such as $\beta$-VAE~\cite{beta-VAE} and FactorVAE~\cite{FactorVAE} encourage axis-aligned independence by strengthening a Kullback–Leibler term, while adversarial approaches (InfoGAN~\cite{chen2016infogan}, DANN~\cite{Ganin2016DNN}) promote domain invariance via discriminators. Post-hoc strategies on frozen encoders include CORAL, which whitens source features and recolors them to match target covariance in a single affine step; its linear variant preserves simplicity but effectively collapses scanner variation into one direction~\cite{SUN2016Coral}. 
PISCO~\cite{pisco-ngweta23a} similarly extracts a single technical direction using a logistic classifier. Its effectiveness has been demonstrated primarily on synthetic augmentations. More recent \emph{contrastive} approaches split latent factors into common- and salient subspaces—contrastive VAE~\cite{abid2019cvae} under a background–target setup, and SepCLR~\cite{louiset2024sepclr} via InfoMax-style contrastive learning without a decoder. While effective, these approaches rely on end-to-end optimization with large-batch contrastive objectives and careful dataset partitioning, which is challenging in multi-scanner, unlabeled clinical data at scale.


\paragraph{Harmonization approaches.}
In clinical CT, harmonization is commonly addressed by: (i) \textit{ComBat}~\cite{Fortin2018ComBat}, which performs feature-wise location/scale adjustment via empirical Bayes; (ii) \textit{adversarial domain adaptation} (e.g., DANN~\cite{Ganin2016DNN}), which learns domain-invariant features jointly with a task network but requires access to target images and careful hyperparameter tuning; and (iii) \textit{feature-matching} such as Deep CORAL~\cite{SUN2016Coral}, which injects a correlation loss into intermediate layers when raw images are available. These methods improve scanner invariance but rarely target the \emph{unsupervised} discovery of tissue types. By disentangling real paired CT patches into technical and biological subspaces using a post-hoc \emph{linear} rotation on a frozen encoder, our approach bridges harmonization and interpretability—aiming for scanner-invariant clustering of anatomy- and disease-related tissue characteristics (e.g., lung boundaries, parenchymal textures, fibrotic patterns) that remain consistent across protocols and support downstream clinical modeling.


\paragraph{Contribution}

This work introduces a post-hoc linear rotation framework that extends PISCO~\cite{pisco-ngweta23a} to disentangle \emph{multiple} acquisition-related technical axes from pre-trained image representations and unlabeled data. The key contributions are as follows: 
(1) we introduce a disentanglement approach that separates real-world variation associated with biology from imaging parameters such as reconstruction kernel, slice thickness, and scanner vendor, representing them in two different sub-spaces of the rotated latent embedding space. This extends prior work to medical imaging data, beyond synthetic augmentations and single style axis constraints. 
(2) The method performs \textit{paired patch sampling} leveraging routine clinical data without requiring disease, segmentation, or task-specific labels.
(3) The approach discovers \textit{stable appearance clusters} across a large number of acquisition settings, derived from biologically meaningful latent representations.
(4) We evaluate the impact of the disentangling on a down-stream disease progression prediction task in an IPF cohort without retraining clusters or any other model parameter.

\section{Method}
\label{sec:method}
\begin{figure}[!t]
    \centering
    \includegraphics[width=1\linewidth]
    {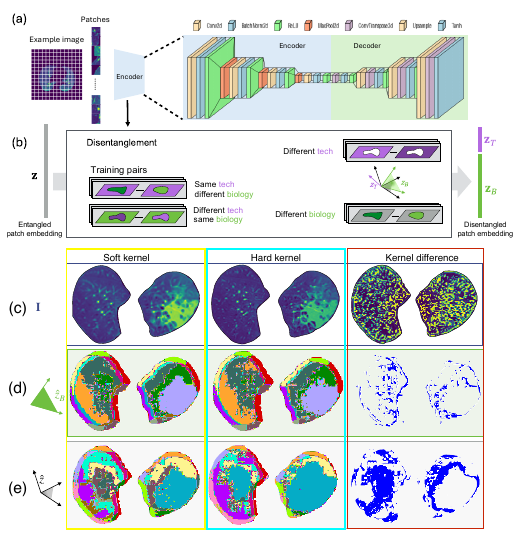}  
    \caption{Method overview: (a) Autoencoder encodes image patches. (b) Paired patches (same anatomy/different imaging parameters or vice versa) guide a rotation separating biology from technical factors. (c) Inputs; (d) clustering in $\mathbf z_B$ (disentangled); (e) clustering in $\mathbf z$ (entangled).}
    \label{fig:method}
\end{figure}

An overview of the approach is shown in Fig.~\ref{fig:method}a-b. We aim to identify biologically meaningful tissue patterns in chest CT that remain stable across heterogeneous acquisition settings of the scanner. From each CT image $\mathbf{I}_i$, square 
patches $\mathbf{p}_i \in \mathbb{R}^{16 \times 16}$  are extracted within a lung mask $\mathbf{L}_i \subseteq \mathbf{I}_i$. These patches are encoded to latent representations $\mathbf{z} \in \mathbb{R}^m$ using a pretrained autoencoder. A learned linear rotation then separates the latent space into two components: (i) a \textbf{biological subspace} ($\mathbf{z}_B$) encoding anatomical or tissue-related structure, and (ii) a \textbf{technical subspace} ($\mathbf{z}_T$) capturing scanner and protocol variation. Clustering in $\mathbf{z}_B$ yields more stable tissue categories across acquisition settings and enhances survival prediction in external cohorts, compared to clustering in the entire latent space.

\subsubsection{Auto-encoder Pre-training:}
\label{sec:CAE}
The encoder \( f_\phi \) is trained using a mean-squared reconstruction loss on image patches. After training, encoder weights are frozen, and the decoder is discarded for all subsequent steps.  

\subsubsection{Sampling Paired Examples for Disentanglement:}
\label{sec:pairs}
To disentangle technical variability from biology, we construct paired latent differences $\Delta \mathbf{z}_{ij}=\mathbf{z}_i-\mathbf{z}_j$ with binary labels: $y_{ij}=1$ if the pair shares anatomical location but differs in scanner/protocol, and $y_{ij}=0$ if the pair shares scanner/protocol but differs in anatomical location. These pairs are drawn from routine clinical data without requiring any disease labels. To ensure meaningful and diverse pairs, we employ FAISS-based hard-negative mining~\cite{johnson2019FAISS}. For each $\ell_2$-normalized embedding $\mathbf{z}_i$, we retrieve its nearest neighbors by cosine similarity and uniformly sample a negative from ranks 10--50. This avoids trivial matches (self or very close duplicates) while injecting stochasticity, since patches are not ordered by patient identity. This yields a training set $\{(\Delta z_{ij},y_{ij})\}$ that forms the basis for isolating technical from biological variation in the latent space.





\subsubsection{Post-hoc Linear disentanglement via SVD rotation:}
We train a \textit{linear logistic classifier} $h_\theta$ on $\{(\Delta z_{ij},y_{ij})\}$ to predict $y_{ij}$ from paired latent differences $\Delta z_{ij}$. After training, the classifier’s weight vector $w\in\mathbb{R}^m$ points in the direction that best separates technical from biological differences. We normalize this vector to obtain the \emph{dominant technical axis} $v_{\text{base}}=w/\lVert w\rVert$ (equivalently obtainable via SVD of the weight matrix when using a linear layer). Next, we project $\Delta z_{ij}$ onto the orthogonal complement of $v_{\text{base}}$ and apply PCA to these residuals to obtain $(r-2)$ additional, mutually orthogonal technical directions $V_{\text{res}}=\{v_2,\ldots,v_{r-1}\}$, with $r$ denoting the dimensionality of the technical subspace $\mathbf{z}_T$ ($1<r<m$).  We also estimate a global mean protocol offset $v_{\text{diff}}$ as the average latent difference across matched acquisitions, capturing any consistent shift between protocols. Stacking these directions and orthonormalizing (QR) yields the projection matrix $P\in\mathbb{R}^{m\times r}$:
\begin{equation}\tag{3}
\label{eq:svd-rot}
\begin{array}{c@{\qquad}c@{\qquad}c}
P=\operatorname{QR}\big[v_{\text{base}},V_{\text{res}},v_{\text{diff}}\big] &
z_T=PP^\top z &
z_B=(I-PP^\top)z
\end{array}
\end{equation}
so that $\mathbf z_T$ carries technical variation, while $\mathbf z_B$ resides in the complementary (biological) subspace used for clustering and downstream analyses.


\subsubsection{Identification of Biological Tissue Types:}
\label{sec:cluster}
After removing the first r technical dimensions, we perform PCA on the residual (“biological”) features $\mathbf z_B$. A k-means clustering algorithm $(k\in\{5,\dots,50\})$ then partitions the embedding into recurrent tissue-appearance patterns $\mathbf{C}_i$. Each patch is assigned by its nearest cluster centroid, forming discrete tissue categories. These cluster labels define tissue cluster profiles for assessing biological consistency across reconstruction kernels. The continuous embeddings $\mathbf z_B$ (or patient-level histograms of their cluster counts) are used as inputs to lightweight predictive models — such as linear, logistic, or Cox regression — for downstream clinical tasks like survival analysis.

\section{Experimental Setup}
\label{sec:setup}
The data encompasses the discovery cohort and an external evaluation cohort. \textit{Discovery cohort:} The in-house set comprises 108 chest-CT patients (seven disease categories) acquired on a \emph{single} vendor and reconstructed with both \textit{soft} and \textit{hard} kernels. After lung segmentation and resampling to \(224{\times}224\) pixel size, 100 co-registered \(16{\times}16\) parenchyma patches are sampled from every matched kernel pair, yielding $\sim5.25\times10^{6}$ patches in total (patient patch counts vary with the number of scans). A patient-level split allocates 36 patients ($\approx$3M patch pairs) to auto-encoder pre-training, 48 patients ($\approx$1.5 M; 20 \% of them for validation) to disentanglement learning, and 24 patients ($\approx$0.75 M) to testing and stability analysis. \textit{External cohort:} External generalization and utility for prediction is evaluated on the OSIC ~\cite{Walsh2024-hp} idiopathic-pulmonary-fibrosis cohort comprising 102 patients acquired on scanners of 4 vendors, with 23 kernels and 15 slice-thickness values. This replication is performed without retraining the encoder disentanglement or clustering, only cluster profiles are extracted and then used for prediction.

\subsection{Evaluation Protocol}\label{sec:Evaluation}
\textit{Cluster stability:} We evaluate the stability of the identified tissue clusters of different imaging parameters. We perform $k$-means (\(k\!=\!5\dots50\))  clustering in two embeddings to identify clusters: \(\mathbf z_B\) or \(\mathbf z\). We evaluate the stability of clusters in the same lung, but imaged with different kernels by Adjusted Rand Index (ARI)~\cite{ARI}, Normalized Mutual Information (NMI)~\cite{NMI}, and the Dice score between the resulting cluster maps. We expect an improvement of stability after disentanglement (\(\mathbf z_B\)). 
\textit{Technical subspace classification:} To validate that scanner factors concentrate in the technical subspace, we train a regularized logistic classifier to predict reconstruction kernel (soft vs.\ hard) from the technical embedding (\(\mathbf z_T\)) and entangled embedding (\(\mathbf z\)). 
\textit{Clinical outcome prediction:} To assess clinical utility of the approach, we evaluate the prediction of outcome based on lung tissue composition profiles on a multi-center cohort of patients comprising substantial imaging parameters heterogeneity. For each patient, the profile of relative cluster volumes \(\sum\mathcal{C}_i\) is used as predictor in  a Cox proportional-hazards model predicting survival. We assess the Hazard ratios (HR) on this external OSIC IPF cohort to compare prognostic power of disentangled  (\(\mathbf z_B\)) and entangled  (\(\mathbf z\)) representations.

\subsection{Implementation Details}
\paragraph{Auto-encoder.} A four-layer convolutional encoder (channels [16,32,64,128] with batch normalization, stride-2 max-pooling) produces a 2×2×128 latent tensor (m=512). Training uses Adam (lr $1\times10^{-4}$, effective batch 128 via gradient accumulation) for 250 epochs with early stopping after 54 epochs (final MSE \( 2.7 \times 10^{-4} \)) and cosine annealing scheduler.

\paragraph{Disentanglement.}  
From the disentanglement split, 15000 slice pairs are randomly chosen
for learning the rotation.  Hard-negative mining is performed with a FAISS~\cite{johnson2019FAISS} GPU index: all latent representations are $\ell_2$-normalised and
each $\mathbf z_i$ retrieves 51 neighbours, with one index uniformly sampled from ranks 10–50.  The resulting $\Delta
\mathbf z$ vectors train a \emph{linear} technical classifier for 50 epochs
(Adam, lr $1\times10^{-3}$, batch 4096, BCE loss).  Singular-value
decomposition yields the dominant style axis; residual technical directions
($r{-}2$) come from PCA on the orthogonal residual, and the global
kernel–difference vector closes the $(r{=}10,30,50)$-dimensional technical
sub-space.

\paragraph{Baseline approaches for comparative evaluation}
The same frozen embeddings, clustering routine, and evaluation metrics are used for all methods.
\textit{(i)DANN}~\cite{Ganin2016DNN}: two-layer encoder (512\(\rightarrow\)512, ReLU) with a three-layer domain classifier (hidden 128). Trained 50 epochs (Adam, LR \(1\!\times\!10^{-4}\), batch 2\,048); gradient-reversal weight \(\lambda = 2/(1+\mathrm{e}^{-10p})-1\). 
\textit{(ii)Deep CORAL}~\cite{SUN2016Coral}: a linear projection (512\(\rightarrow\)512) minimises the CORAL loss for 50 epochs (Adam, LR \(1\!\times\!10^{-3}\)); mean-shift correction is applied at inference. 
\textit{(iii)$\beta$-VAE}~\cite{beta-VAE}: convolutional $\beta$-VAE (latent 64) trained 20 epochs on 100\,000 randomly sampled patches (Adam, LR \(1\!\times\!10^{-4}\), batch 64) with \(\beta{=}0.05\).  The 50\,\% most stable latent dimensions (lowest inter-domain variance) are retained. 
\textit{(iv)ComBat}~\cite{Fortin2018ComBat}: empirical-Bayes harmonization with kernel as batch co-variate and anatomical location ID as continuous co-variate.  Estimates are fit on 20\,\% of the training embeddings and applied in 50\,000-sample chunks.

\paragraph{Clustering and prediction.}  We cluster the biological embedding (\(\mathbf z_B\)) with $k$-means ($k\!=\!5\text{--}50$) using Euclidean and cosine distances, selecting the better-balanced solution. For each patient, the fraction of lung volume per cluster defines a \emph{cluster-volume profile} used as covariates in a Cox proportional-hazards model.
Centroids learned on the discovery cohort are applied unchanged to assign each lung voxel in the external IPF cohort to one tissue type, resulting in a labeling of the entire lung volume and the construction of the corresponding profile. 

\section{Results}\label{sec:results}

\begin{figure}[t]
\centering
\includegraphics[width=0.95\textwidth]{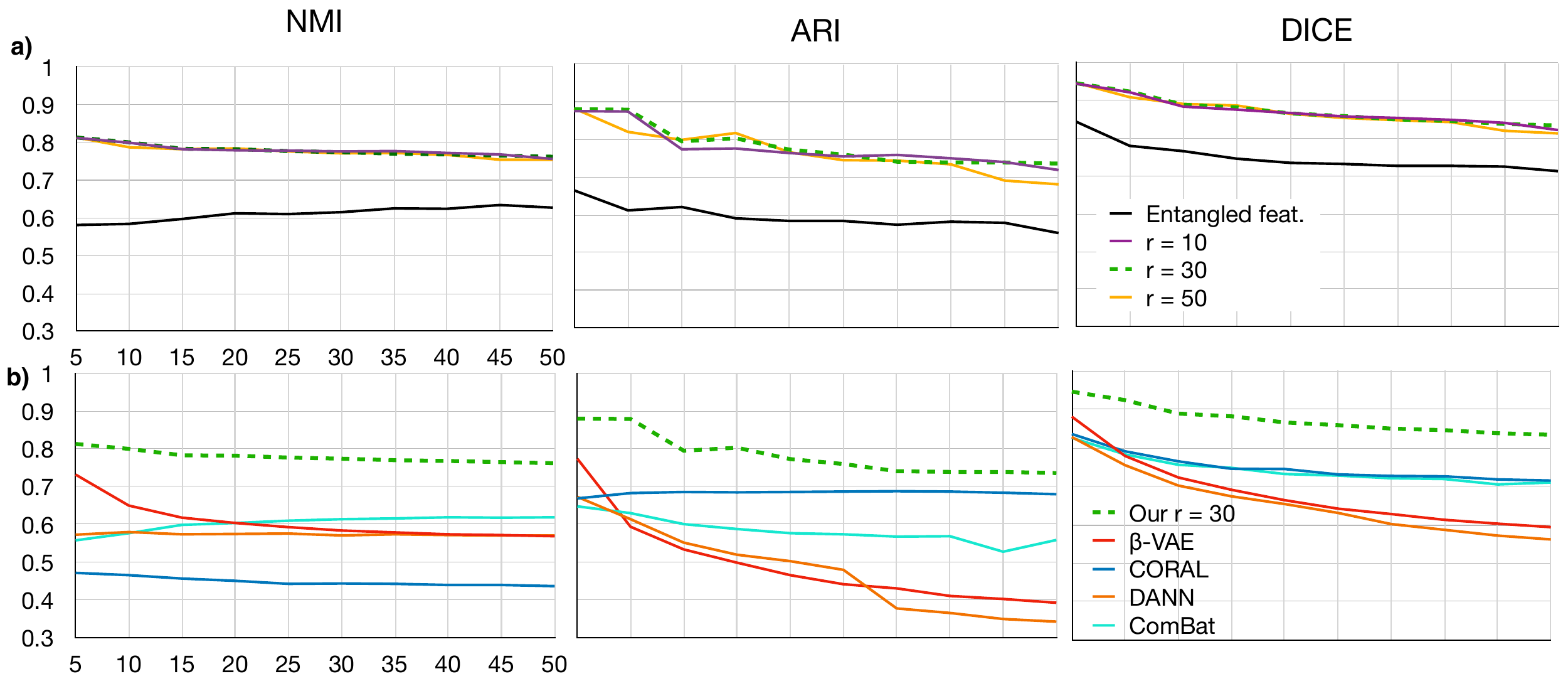}
\caption{Clustering stability vs.\ number of clusters ($k{=}5\text{--}50$). ARI/NMI/Dice for (a) biological embedding $\mathbf z_B$ at $r\!\in\!\{10,30,50\}$ vs.\ entangled $\mathbf z$; (b) $\mathbf z_B$ ($r{=}30$) compared with DANN, Deep CORAL, $\beta$-VAE, and ComBat (ours dashed).
}
\label{fig:ari_nmi_score}
\end{figure}

\subsubsection{Cluster stability is higher for the disentangled representation}
\label{sec:results:stability}
Across $k\!=\!5$–$50$, biological embeddings ($\mathbf z_B$) consistently outperform entangled embeddings ($\mathbf z$) in ARI/NMI/Dice (Fig.~\ref{fig:ari_nmi_score}a), with mean gains of $+19.0\%$ (ARI), $+16.9\%$ (MNI), and $+12.4\%$ (Dice) across all values for $r$. The best $r$ depends on the granularity: $r\!=\!50$ for coarse partitions ($k\!\le\!15$) and $r\!=\!30$ for finer ones ($k\!\ge\!35$). Under the same protocol, $\mathbf z_B$ ($r\!=\!30$) also surpasses DANN, Deep CORAL, $\beta$-VAE, and ComBat (Fig.~\ref{fig:ari_nmi_score}b); Deep CORAL is closest (still $8$–$13$\% lower), while ComBat performs worst.

\subsubsection{Imaging parameters are represented primarily in the technical sub-space:}
To validate the technical subspace, we trained a logistic classifier to predict the reconstruction kernel (soft vs.\ hard) from technical embeddings ($\mathbf z_T$,$r\!=\!30$) and compared it to entangled embeddings ($\mathbf z$). Accuracies are summarized in Table~\ref{tab:kernel_clf}; the accuracy for $\mathbf z_T$ is substantially higher for  $\mathbf z$, indicating scanner variation is encoded in the technical subspace, with high separability.

\subsubsection{Qualitative results}
Figure~\ref{fig:method}(c–e) shows cluster maps for the same axial slice under $C_{soft}$ (yellow) and $C_{hard}$ (cyan) kernels, with a red binary map highlighting voxels where assignments differ across kernels ($C_{\mathrm{soft}}\neq C_{\mathrm{hard}}$). Clustering in the biological embedding $\mathbf z_B$ (Fig.~\ref{fig:method}d) remains anatomically coherent and largely invariant to kernel changes (e.g., lung borders, fibrotic reticulation), whereas clustering from the entangled embedding $\mathbf z$ (Fig.~\ref{fig:method}e) shifts with the difference map, indicating sensitivity to kernel artifacts.

\subsubsection{Clinical outcome prediction}
Patient-level cluster-volume profiles were computed on the external OSIC IPF cohort and used as covariates in a Cox proportional-hazards model (no retraining of embedding, disentanglement, or clusters). As summarized in Table~\ref{tab:cox_hr}, profiles from the biological embedding $z_B$ yield higher hazard ratios (HR) than the entangled embedding $z$ for most cluster granularities $k\in\{10,20,30,40,50\}$. The only exception is $k{=}40$, where over-fragmentation produces small, unstable clusters and lowers HR; increasing to $k{=}50$ recovers and surpasses prior performance (HR $=3.40$).

\begin{table*}[t]
\centering
\caption{(a) Kernel classification accuracy (soft vs.\ hard) using $\mathbf z_T$ vs.\ $\mathbf z$; (b) OSIC IPF: Cox HRs from patient cluster profiles ($\mathbf z_B$ vs.\ $\mathbf z$) across $k$.}
\label{tab:cox_and_kernel}
\setlength{\tabcolsep}{4pt}
\begin{subtable}[t]{0.5\textwidth}
\centering
\scriptsize
\caption{Kernel classification accuracy.}
\label{tab:kernel_clf}
\begin{tabular}{|l|c|c|r|}
\hline
\textbf{Features} & \textbf{Soft} & \textbf{Hard} & \textbf{Overall} \\
\hline
Technical ($\mathbf z_T$) & \textbf{0.865} & \textbf{0.966} & \textbf{0.916} \\
Entangled ($\mathbf z$)  & 0.620          & 0.608          & 0.614 \\
\hline
\end{tabular}
\end{subtable}\hfill
\begin{subtable}[t]{0.5\textwidth}
\centering
\scriptsize
\caption{Cox HRs across $k\in\{10,\dots,50\}$.}
\label{tab:cox_hr}
\begin{tabular}{|l|c|c|c|c|r|}
\hline
\textbf{k} & \textbf{10} & \textbf{20} & \textbf{30} & \textbf{40} & \textbf{50} \\
\hline
HR$_{\mathbf z_B}$ & \textbf{2.59} & \textbf{2.22} & \textbf{2.85} & 1.52 & \textbf{3.4} \\
HR$_{\mathbf z}$   & 2.19          & 1.97          & 1.96          & \textbf{2.17} & 2.36 \\
\hline
\end{tabular}
\end{subtable}
\end{table*}

\section{Discussion and Conclusion}\label{sec:conclusion}

This study presents a lightweight post-hoc approach that disentangles biological appearance from technical factors. The method operates on a frozen auto-encoder and requires neither decoder retraining nor disease-specific supervision. The results show a consistent increase in cluster stability compared to entangled representations and the ability to generalize representations and tissue clusters without adaptation to a different unseen, multi-centre IPF cohort. On this cohort, the tissue profiles resulting from disentangled representations improve survival prediction for high cluster numbers. Compared to established harmonization techniques such as DANN, Deep CORAL, $\beta$-VAE, and ComBat, our disentangled embeddings achieve higher cluster stability, highlighting the effectiveness of linear post-hoc separation in capturing biologically meaningful structure.

\paragraph{Limitations.}
The method assumes paired reconstructions from the same patient. The internal cohort includes only two scanners per vendor, and rare protocol combinations may require re-estimating or enlarging the technical subspace. We did not benchmark contrastive VAE~\cite{abid2019cvae}, or SepCLR~\cite{louiset2024sepclr} as they require end-to-end, large-batch contrastive training with explicit background-target splits-impractical for multi-scanner, unlabeled cohorts and misaligned with our frozen-encoder, post-hoc design. Finally, although the sample size is modest, evaluation on an independent, heterogeneous external cohort supports robustness.

\paragraph{Conclusion.}
Disentangling variability associated with biology and image acquisition is feasible on routine imaging data. It markedly improves the stability of tissue appearance clusters obtained by unsupervised machine learning, and improves clinical utility in downstream tasks such as prediction based on  tissue profiles. The proposed rotation approach offers a practical step towards reliable, label-free biomarker discovery in heterogeneous real-world imaging data.

\paragraph{Prospect of application.} 
Using routine paired scans to learn disentangling variability, the method generalizes beyond training and suits retrospective multi-centre studies; integrated into trial pipelines, its lightweight rotation can harmonize CT and enable scanner-invariant tissue profiling for disease prognosis.

\begin{credits}
\subsubsection{\ackname} This research has been partially funded by the Austrian Science Fund (FWF, P35189 - ONSET), by the European Commission (No. 101080302 - AI-POD), the International Atomic Energy Agency IAEA (ZODIAC project) and the Open Source Imaging Consortium (OSIC, https://www.osicild.org).

\subsubsection{\discintname}
Georg Langs is shareholder and co-founder of Contextflow GmbH. Other authors have no competing interests to declare.
\end{credits}

\bibliographystyle{splncs04}

\begin{thebibliography}{10}
\providecommand{\url}[1]{\texttt{#1}}
\providecommand{\urlprefix}{URL }
\providecommand{\doi}[1]{https://doi.org/#1}

\bibitem{abid2019cvae}
Abid, A., Zou, J.Y.: Contrastive variational autoencoder enhances salient features. CoRR  \textbf{abs/1902.04601} (2019), \url{http://arxiv.org/abs/1902.04601}

\bibitem{Boedeker2004-jx}
Boedeker, K.L., McNitt-Gray, M.F., Rogers, S.R., Truong, D.A., Brown, M.S., Gjertson, D.W., Goldin, J.G.: Emphysema: effect of reconstruction algorithm on {CT} imaging measures. Radiology  \textbf{232}(1),  295--301 (Jul 2004)

\bibitem{chen2016infogan}
Chen, X., Duan, Y., Houthooft, R., Schulman, J., Sutskever, I., Abbeel, P.: Infogan: Interpretable representation learning by information maximizing generative adversarial nets (2016), \url{http://arxiv.org/abs/1606.03657}, cite arxiv:1606.03657

\bibitem{Fisher2004-tz}
Fisher, R.I., Miller, T.P., O'Connor, O.A.: Diffuse aggressive lymphoma. Hematology Am. Soc. Hematol. Educ. Program  \textbf{2004}(1),  221--236 (2004)

\bibitem{Fortin2018ComBat}
Fortin, J.P., Cullen, N., Sheline, Y.I., Taylor, W.D., Aselcioglu, I., Cook, P.A., Adams, P., Cooper, C., Fava, M., McGrath, P.J., McInnis, M., Phillips, M.L., Trivedi, M.H., Weissman, M.M., Shinohara, R.T.: Harmonization of cortical thickness measurements across scanners and sites. NeuroImage  \textbf{167},  104--120 (2018). \doi{https://doi.org/10.1016/j.neuroimage.2017.11.024}, \url{https://www.sciencedirect.com/science/article/pii/S105381191730931X}

\bibitem{Ganin2016DNN}
Ganin, Y., Ustinova, E., Ajakan, H., Germain, P., Larochelle, H., Laviolette, F., Marchand, M., Lempitsky, V.: Domain-adversarial training of neural networks (2016), \url{https://arxiv.org/abs/1505.07818}

\bibitem{Gierada2010-ik}
Gierada, D.S., Bierhals, A.J., Choong, C.K., Bartel, S.T., Ritter, J.H., Das, N.A., Hong, C., Pilgram, T.K., Bae, K.T., Whiting, B.R., Woods, J.C., Hogg, J.C., Lutey, B.A., Battafarano, R.J., Cooper, J.D., Meyers, B.F., Patterson, G.A.: Effects of {CT} section thickness and reconstruction kernel on emphysema quantification relationship to the magnitude of the {CT} emphysema index. Acad. Radiol.  \textbf{17}(2),  146--156 (Feb 2010)

\bibitem{beta-VAE}
Higgins, I., Matthey, L., Pal, A., Burgess, C., Glorot, X., Botvinick, M., Mohamed, S., Lerchner, A.: beta-{VAE}: Learning basic visual concepts with a constrained variational framework. In: International Conference on Learning Representations (2017), \url{https://openreview.net/forum?id=Sy2fzU9gl}

\bibitem{ARI}
Hubert, L., Arabie, P.: Comparing partitions. J. Classif.  \textbf{2}(1),  193--218 (Dec 1985)

\bibitem{johnson2019FAISS}
Johnson, J., Douze, M., J{\'e}gou, H.: Billion-scale similarity search with {GPUs}. IEEE Transactions on Big Data  \textbf{7}(3),  535--547 (2019)

\bibitem{FactorVAE}
Kim, H., Mnih, A.: Disentangling by factorising. In: Dy, J., Krause, A. (eds.) Proceedings of the 35th International Conference on Machine Learning. Proceedings of Machine Learning Research, vol.~80, pp. 2649--2658. PMLR (10--15 Jul 2018), \url{https://proceedings.mlr.press/v80/kim18b.html}

\bibitem{Lee2024-nu}
Lee, S., Kim, S., Koh, G., Ahn, H.: Identification of time-series pattern marker in its application to mortality analysis of pneumonia patients in intensive care unit. J. Pers. Med.  \textbf{14}(8), ~812 (Jul 2024)

\bibitem{louiset2024sepclr}
Louiset, R., Duchesnay, E., Grigis, A., Gori, P.: Separating common from salient patterns with contrastive representation learning (2024), \url{https://arxiv.org/abs/2402.11928}

\bibitem{Mets2012-ao}
Mets, O.M., de~Jong, P.A., van Ginneken, B., Gietema, H.A., Lammers, J.W.J.: Quantitative computed tomography in {COPD}: possibilities and limitations. Lung  \textbf{190}(2),  133--145 (Apr 2012)

\bibitem{pisco-ngweta23a}
Ngweta, L., Maity, S., Gittens, A., Sun, Y., Yurochkin, M.: Simple disentanglement of style and content in visual representations. In: Krause, A., Brunskill, E., Cho, K., Engelhardt, B., Sabato, S., Scarlett, J. (eds.) Proceedings of the 40th International Conference on Machine Learning. Proceedings of Machine Learning Research, vol.~202, pp. 26063--26086. PMLR (23--29 Jul 2023), \url{https://proceedings.mlr.press/v202/ngweta23a.html}

\bibitem{Pan2023-qy}
Pan, J., Hofmanninger, J., Nenning, K.H., Prayer, F., R{\"o}hrich, S., Sverzellati, N., Poletti, V., Tomassetti, S., Weber, M., Prosch, H., Langs, G.: Unsupervised machine learning identifies predictive progression markers of {IPF}. Eur. Radiol.  \textbf{33}(2),  925--935 (Feb 2023)

\bibitem{Prayer2021-qv}
Prayer, F., Hofmanninger, J., Weber, M., Kifjak, D., Willenpart, A., Pan, J., R{\"o}hrich, S., Langs, G., Prosch, H.: Variability of computed tomography radiomics features of fibrosing interstitial lung disease: A test-retest study. Methods  \textbf{188},  98--104 (Apr 2021)

\bibitem{NMI}
Strehl, A., Ghosh, J.: Cluster ensembles --- a knowledge reuse framework for combining multiple partitions. J. Mach. Learn. Res.  \textbf{3}(null),  583–617 (Mar 2003). \doi{10.1162/153244303321897735}, \url{https://doi.org/10.1162/153244303321897735}

\bibitem{SUN2016Coral}
Sun, B., Saenko, K.: Deep coral: Correlation alignment for deep domain adaptation. In: Hua, G., J{\'e}gou, H. (eds.) Computer Vision -- ECCV 2016 Workshops. pp. 443--450. Springer International Publishing, Cham (2016)

\bibitem{Walsh2024-hp}
Walsh, S.L.F., De~Backer, J., Prosch, H., Langs, G., Calandriello, L., Cottin, V., Brown, K.K., Inoue, Y., Tzilas, V., Estes, E., {Open Source Imaging Consortium (OSIC)}: Towards the adoption of quantitative computed tomography in the management of interstitial lung disease. Eur. Respir. Rev.  \textbf{33}(171) (Jan 2024)

\bibitem{Yoon2024}
Yoon, J.S., Oh, K., Shin, Y., Mazurowski, M.A., Suk, H.I.: Domain generalization for medical image analysis: A review. Proceedings of the IEEE  \textbf{112}(10),  1583--1609 (2024). \doi{10.1109/JPROC.2024.3507831}

\bibitem{Zhou2021}
Zhou, S.K., Greenspan, H., Davatzikos, C., Duncan, J.S., Van~Ginneken, B., Madabhushi, A., Prince, J.L., Rueckert, D., Summers, R.M.: A review of deep learning in medical imaging: Imaging traits, technology trends, case studies with progress highlights, and future promises. Proceedings of the IEEE  \textbf{109}(5),  820--838 (2021). \doi{10.1109/JPROC.2021.3054390}

\end{thebibliography}

\end{document}